\documentclass[10pt,twocolumn,letterpaper]{article}

\usepackage{cvpr}              

\usepackage{microtype}
\usepackage{xspace}
\usepackage{makecell}

\usepackage{multirow}
\usepackage{wrapfig}
\usepackage[accsupp]{axessibility}  

\definecolor{cvprblue}{rgb}{0.21,0.49,0.74}
\usepackage[pagebackref,breaklinks,colorlinks,allcolors=cvprblue]{hyperref}
\usepackage[ruled,vlined]{algorithm2e}
\usepackage{algpseudocode}

\def\paperID{5191} 
\def\confName{CVPR}
\def\confYear{2025}

\title{Doppelgangers++: Improved Visual Disambiguation with Geometric 3D Features}

\author{
Yuanbo Xiangli$^1$
\quad 
Ruojin Cai$^1$
\quad
Hanyu Chen$^1$ 
\quad
Jeffrey Byrne$^2$ 
\quad
Noah Snavely$^1$ \\
$^1$Cornell University, $^2$Visym Labs 
}

\begin{document}

\twocolumn[{%
	\renewcommand\twocolumn[1][]{#1}%
	\maketitle
	\begin{center}
	    \vspace*{-20pt}
		\centering
		\includegraphics[width=.98\textwidth]{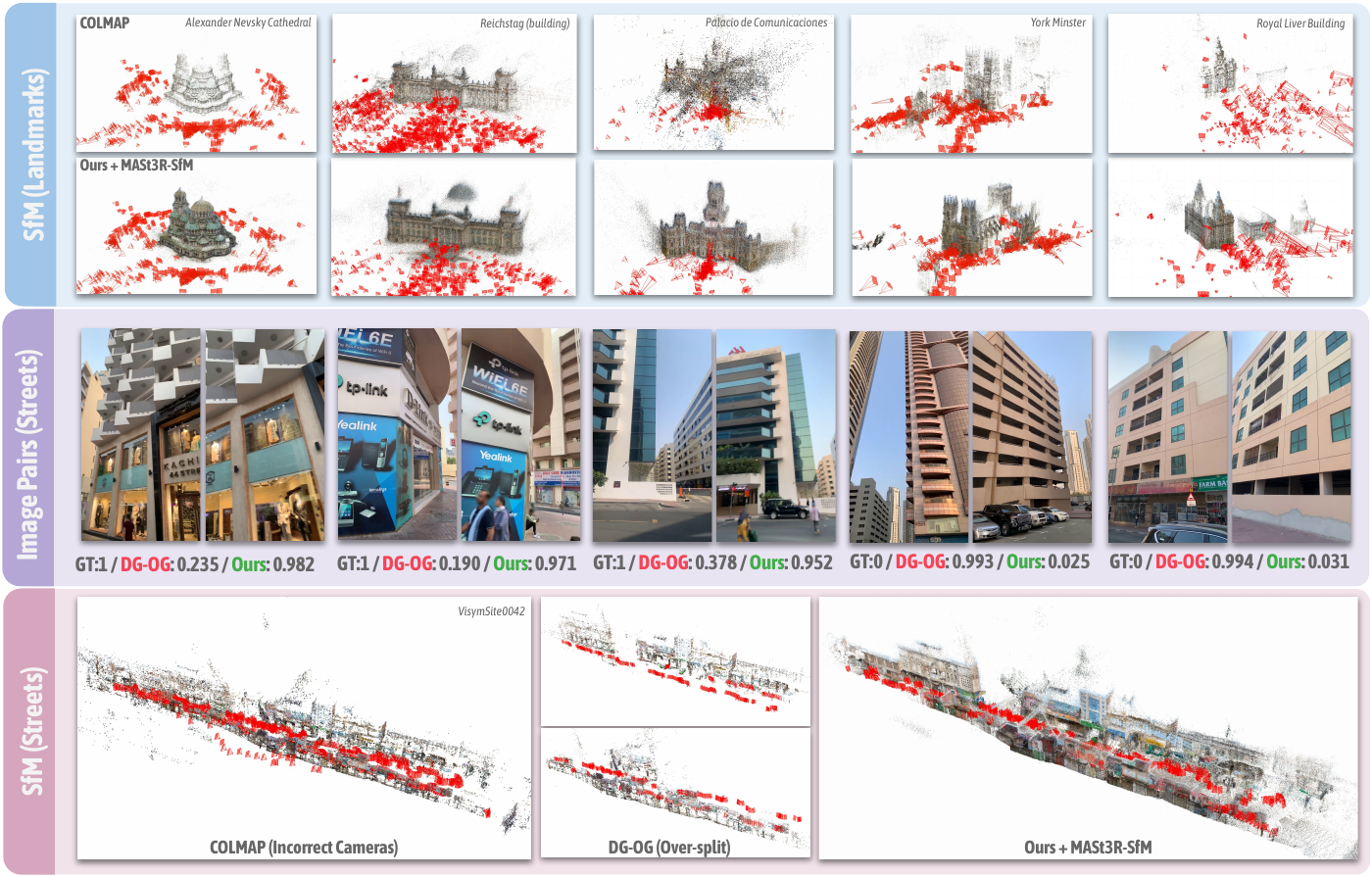}
		\vspace*{-5pt}
    	\captionof{figure}{
        \footnotesize
        Visual aliasing, or doppelgangers, poses severe challenges to 3D reconstruction. 
        We propose \emph{Doppelganger++}, an enhanced pairwise image classifier that excels in visual disambiguation across diverse and challenging scenes.
        \textbf{(Top)} We seamlessly integrate Doppelganger++ into SfM, successfully disambiguating each scene.
        \textbf{(Middle)} Compared to prior work (which we refer to as DG-OG~\cite{Cai2023DoppelgangersLT}), Doppelgangers++ is more robust for everyday scenes, showing improved accuracy and robustness. We show pairs that DG-OG classifies incorrectly and ours gets correct. 
        \textbf{(Bottom)} Our new \emph{VisymScenes} dataset, featuring complex daily scenes, is particularly challenging for COLMAP and DG-OG, but our method can achieve correct and complete reconstructions.}
		\label{fig:teaser}
	\end{center}%
}]




\newcommand{\bb}[1]{\textcolor{blue!80!black}{\underline{#1}}}

\newcommand{\misscite}{\textcolor{red}{[C]}}
\newcommand{\missref}{\textcolor{red}{[R]}}
\newcommand{\amber}[1]{\textcolor{orange}{{[#1]}}}
\newcommand{\mean}{\mathop{\mathrm{mean}}}
\newcommand{\visym}{VisymScenes\xspace}
\newcommand{\dgdata}{DG\xspace}

\definecolor{darklavender}{rgb}{0.45, 0.31, 0.59}
\definecolor{darkviolet}{rgb}{0.58, 0.0, 0.83}

\definecolor{visible-blue}{rgb}{0.286, 0.525, 0.910}

\definecolor{tabfirst}{rgb}{1, 0.7, 0.7} 
\definecolor{tabsecond}{rgb}{1, 0.85, 0.7} 
\definecolor{tabthird}{rgb}{1, 1, 0.7} 

\definecolor{placeholder}{rgb}{0.6,0.8,0.95}
\newcommand{\placeholder}[1]{\textcolor{placeholder}
{\rule{\linewidth}{#1}}}

\begin{abstract}

Accurate 3D reconstruction is frequently hindered by visual aliasing, where visually similar but distinct surfaces (aka, \emph{doppelgangers}), are incorrectly matched.
These spurious matches distort the structure-from-motion (SfM) process, leading to misplaced model elements and reduced accuracy. 
Prior efforts addressed this with CNN classifiers trained on curated datasets, but these approaches struggle to generalize across diverse real-world scenes and can require extensive parameter tuning.
In this work, we present \emph{Doppelgangers++}, a method to enhance doppelganger detection and improve 3D reconstruction accuracy. 
Our contributions include a diversified training dataset that incorporates geo-tagged images from everyday scenes
to expand robustness beyond landmark-based datasets. 
We further propose a Transformer-based classifier that leverages 3D-aware features from the MASt3R model, achieving superior precision and recall across both in-domain and out-of-domain tests. 
Doppelgangers++ integrates seamlessly into standard SfM and MASt3R-SfM pipelines, offering efficiency and adaptability across varied scenes. 
To evaluate SfM accuracy, we introduce an automated, geotag-based method for validating reconstructed models, eliminating the need for manual inspection. Through extensive experiments, we demonstrate that Doppelgangers++ significantly enhances pairwise visual disambiguation and improves 3D reconstruction quality in complex and diverse scenarios.

\end{abstract}
    
\section{Introduction}
\label{sec:intro}
Visual aliasing---confusing two surfaces that look the same but are nonetheless distinct---is an pernicious problem in 3D reconstruction and SLAM systems. 
Pairs of images that depict visually similar yet distinct 3D surfaces (called \emph{doppelgangers} by Cai \etal~\cite{Cai2023DoppelgangersLT}) can generate spurious correspondence at the feature matching stage of 3D reconstruction, leading to erroneous downstream reconstructions that feature distorted geometry or incorrectly fused elements.
Therefore, to ensure the accuracy of 3D reconstruction, it is critical to distinguish truly matching images from illusory matches arising from doppelganger pairs.

In recent work, this visual disambiguation problem was formulated as a binary classification task on image pairs~\cite{Cai2023DoppelgangersLT}.
The authors collected a Internet dataset of visually similar doppelganger pairs (as well as truly matching images) from \href{https://commons.wikimedia.org/wiki/Main_Page}{Wikimedia Commons}, then trained a CNN to classify image pairs as correct or incorrect matches.
This classifier can be used to take a feature match graph computed from a set of photos, remove incorrect edges between doppelganger image pairs, then reconstruct a correct model using a structure from motion (SfM) pipeline like COLMAP~\cite{schonberger2016structure}.
While that work shows promising improvements on reconstruction problems, we find that it can still be brittle:
First, the 3D reconstruction task demands that the classifier have has precision---even a few bad edges remaining in the image match graph can lead to an incorrect 3D model.
Second, the reconstruction task is quite sensitive to the threshold on the classifier score used to prune edges from the match graph. 
Finally, their method was trained solely on landmark Internet photos and does not reliably generalize to new scenarios, such as more structured captures of everyday scenes, like streets and office buildings.

In this paper, we aim to address these issues and improve doppelganger classification in several key ways: 
\begin{enumerate}
\item We identify ways to expand and diversify doppelgangers training data. In particular, we leverage semi-structured image data with geographic annotations (rough GPS position and orientation), captured from everyday scenes with the Visym Collector platform~\cite{byrne2023fine}.
\item We switch from a CNN-based classifier to leveraging features from MASt3R~\cite{Duisterhof2024MASt3RSfMAF}, a recent transformer-based geometric model that computes point clouds from two input views. 
Specifically, we feed an image pair into a pre-trained MASt3R model, collect the intermediate features decoded by MASt3R, and train a classification head to map these features to a doppelganger score.
\end{enumerate}
We call our method \emph{Doppelgangers++}. 
Our enhanced model achieves higher precision and recall in pairwise classification, 
and generalizes better across a broader range of scenes and capture scenarios.
Doppelganger++ integrates seamlessly into SfM pipelines for 3D reconstruction disambiguation. 
We further propose to leverage geo-tagged map images to quantitatively evaluate the correctness and completeness of reconstructed models,
in comparison to manual inspection as required in previous approaches.
Through extensive experiments we show that our model leads to more accurate and complete reconstruction results,
and is less sensitive to the threshold used for pruning doppelganger matches.

\section{Related Work}
\label{sec:related_work}

\noindent \textbf{Local feature matching and 3D learning.} Local feature matching methods have proven 
effective at establishing correspondences between pairs of images in SfM pipelines. 
Classic methods like COLMAP~\cite{schonberger2016structure} 
rely on the tried-and-true SIFT features~\cite{lowe2004distinctive} to find correspondence.
Modern learning-based, data-driven approaches have improved the quality of local feature matching \cite{Yi2016LIFT, detone2018superpoint,rocco2018neighbourhood, sun2021loftr}.
More recently, the DUSt3R~\cite{Wang2023DUSt3RG3} framework has proven to excel in a variety of 3D reconstruction tasks by 
estimating dense 3D point clouds from pairs of images.
MASt3R~\cite{Leroy2024GroundingIM} was proposed as an extension to DUSt3R that specifically targets the task of predicting dense feature matches between image pairs. 
All of these local feature matching methods are traditionally optimized to maximize the number of correspondences they find between similar image regions. 
While their ability to identify feature matches has greatly improved, they generally lack the ability to incorporate \emph{negative evidence} into their predictions, and so they often find matches between regions that do not actually correspond to the same 3D surface, particularly within doppelganger image pairs.
However, we show that features internal to MASt3R can be repurposed for doppelganger detection.

\medskip
\noindent \textbf{Disambiguation in SfM and image matching.} Disambiguating similar structures and repeated patterns in SfM is a long-standing challenge. Prior work has mainly relied on heuristics-based analysis to detect conflicting relations and ambiguities in the structure of the underlying scene graph~\cite{zach2010disambiguating, wilson2013network, yan2017distinguishing,Kataria2020ImprovingSF,Manam2024LeveragingCT,Shen2016GraphBasedCM,Wang2018HierarchicalIL,Gong2024ACD,Peng2022ViewGC} or among image-level correspondence (or the lack thereof)~\cite{zach2008can, jiang2012seeing, heinly2014correcting, Roberts2011StructureFM,Heinly2014RecoveringCR,Ceylan2014CoupledSA}. 
While such methods have shown some success in detecting and resolving ambiguities in SfM pipelines, one fundamental limitation 
that they do not consider the rich information contained in the images themselves. 
In contrast, 
Doppelgangers~\cite{Cai2023DoppelgangersLT} avoids handcrafted heuristics and takes a data-driven approach to 
the basic problem of identifying doppelganger image pairs. 
They train a CNN that classifies a pair of images as either positive or negative, with LoFTR~\cite{sun2021loftr}-extracted keypoints and match masks passed as auxiliary input. 
During inference, the binary classification results are used to pre-process the scene graph obtained from COLMAP to filter out spurious matches prior to running SfM. 
However, the method's reliance on COLMAP feature extraction and matching and the need for auxiliary mask information from LoFTR
introduces significant overhead and complexity to the overall pipeline, making it difficult to 
scale to larger scenes. In addition, we find that this prior method is brittle: it can fail to generalize to domains beyond landmarks (e.g., to office buildings), and it can require parameter tuning to get good results.

\medskip
\noindent \textbf{Differentiable SfM.} To improve upon traditional SfM, several differentiable SfM pipelines have been proposed to optimize the entire 3D reconstruction process. 
These pipelines often include one or more components that are trained from large 
datasets. MASt3R-SfM~\cite{Duisterhof2024MASt3RSfMAF} proposes an SfM pipeline that uses MASt3R
features for scene graph construction and coarse-to-fine alignment. VGGSfM~\cite{wang2024vggsfm} decomposes the problem into four stages: point tracking, initial camera estimation, triangulation, and bundle adjustment, each of which is 
differentiable and can be optimized in an end-to-end manner. ACE0~\cite{brachmann2025scene} uses a scene coordinate regression network to progressively register new images to a global scene, while the network itself is iteratively trained and refined from the registered images. These methods have shown promising results in optimizing the SfM pipeline, but are not specifically designed to address the doppelganger problem, and are still prone to producing incorrect reconstructions when faced with scenes with repeated patterns or similar structures.

\section{Method}
\label{sec:method}

We refer to a ``doppelganger'' as a case where distinct objects or surfaces look almost identical, leading algorithms to confuse one for the other~\cite{Cai2023DoppelgangersLT}. 
In this work, we aim to improve the doppelganger classifier via expanded and diversified training data~(Sec.~\ref{subsec:visym_dataset}), and leveraging geometric 3D features learned from pairwise reconstruction~\cite{Leroy2024GroundingIM}~(Sec.~\ref{subsec:imrpoved_classifier}).
We further propose an approach to quantitatively evaluate the correctness of SfM results in term of doppelgangers~(Sec.~\ref{subsec:dopp_evaluation}), instead of manual inspection adopted in previous work.

\begin{figure*}
  \centering
   \includegraphics[width=\linewidth]{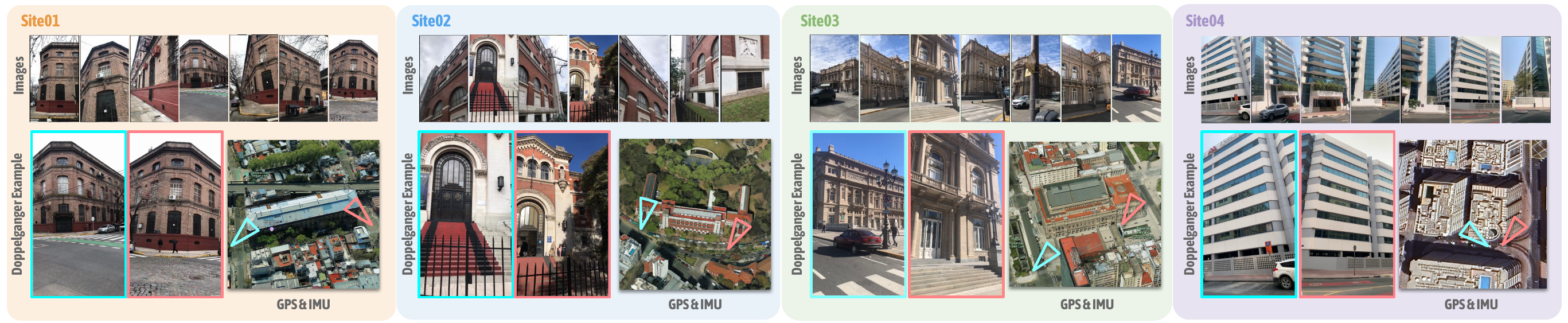}
   \vspace{-15pt}
   \caption{\textbf{\visym examples}. This new dataset includes residential areas, landmarks, historical sites, business districts, and more. Here, we present four example sites. The top row shows subsets of images captured within each site. The bottom row displays pairs of visually similar but geographically distinct images from each site along with their recorded geolocations on a map. These examples demonstrate that doppelganger issues are prevalent in everyday scenes, presenting significant challenges for reliable 3D reconstruction and image matching. }
   \label{fig:visym_dopp_example}
   \vspace{-5pt}
\end{figure*}

\subsection{The \visym dataset}
\label{subsec:visym_dataset}
Cai~\etal~\cite{Cai2023DoppelgangersLT} introduced the Doppelganger dataset, built on global landmarks photos sourced from \href{https://commons.wikimedia.org/wiki/Main_Page}{Wikimedia Commons}, with viewing direction~(\eg, North, South) to identify doppelganger image pairs. 
While the classifier trained on this dataset proved effective, 
we find that it struggles to generalize well to scenes beyond the dataset’s domain.

We enhance and expand their training set to improve the robustness of doppelganger classification, by incorporating 
casually captured images from a wider range of scenes.
To ensure sufficient diversity in the dataset,
we introduce the \emph{\visym} dataset.
\visym consists of 258K ground images with GPS/IMU metadata,
recorded at 149 sites in 42 cities and 15 countries, collected with the Visym Collector Platform~\cite{byrne2023fine} (details in supplementary). 
Each image is accompanied by metadata, such as GPS coordinates, device compass direction, and intrinsic camera calibration. 
While this metadata can be noisy, it still provides valuable information for identifying potential doppelganger pairs. 
For example, if two images exhibit numerous geometrically consistent local feature matches yet were taken from distant locations, this serves as strong evidence that the pair is likely to be a doppelganger. Examples are shown in Fig.~\ref{fig:visym_dopp_example}. 

We develop a series of filtering rules applied to all matched image pairs identified by the COLMAP feature matching
module~\cite{schonberger2016structure}.
Note that these rules are designed according to the capture style of the Visym Collector, 
where cameras consistently focus on the scene of interest and maintain a reasonable distance from it.
For a given pair of images, we use the (metric) distance between the camera centers $r$, angle between their viewing directions $\theta$, and camera intrinsics $\mathcal{K}$, all derived from camera metadata, to identify confident negative (doppelganger) and positive (correct match) pairs.
To identify confident negative pairs,
we first identify spatially distant matching pairs, and classify them as doppelgangers.
Then, we classify the remaining (nearby) pairs 
into three cases:
the intersection point of the viewing directions is 
1) in front of both cameras,
2) behind both cameras;
or 3) in front of one camera and behind the other.
For case 1), we label pairs with very large view angles~(\eg, $>160^\circ$) as negative, since they likely capture different 3D surfaces or mostly non-overlapping content.
For case 2), if their view angle exceeds the diagonal field of view, then their view frustums are unlikely to overlap, so we label them as doppelgangers.
For case 3), we check for frustum overlap using camera intrinsics; if no intersection is detected, the pair is considered a doppelganger.
A similar series of rules are designed for mining positive pairs (\ie, true matches). 
Details of computing view intersection and filtering algorithms for both negative and positive pairs can be found in the supplementary.
With these rules, we mined in total 53K positive and negative pairs across 33 sites for our doppelganger task.

\begin{figure*}[t]
  \centering
   \includegraphics[width=\linewidth]{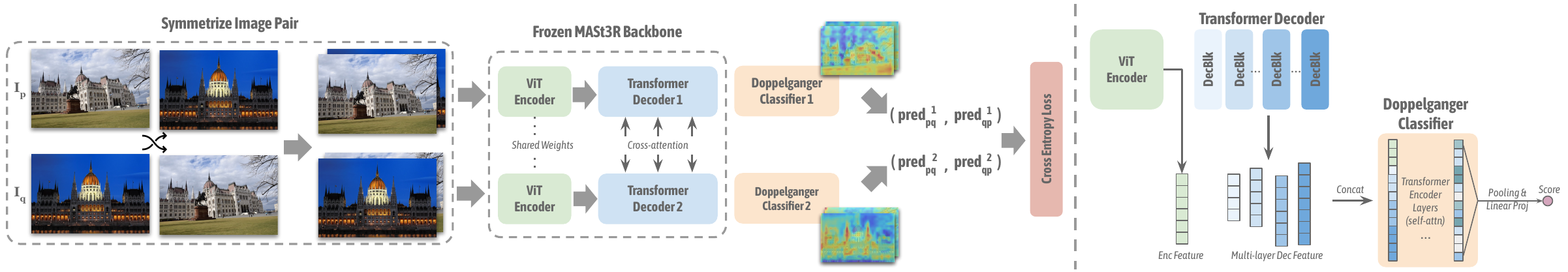}
   \vspace{-20pt}
   \caption{\textbf{Model design.} \textbf{(Left)} Given an image pair, we first create a symmetrized version of the pair and feed it into the frozen MASt3R model. 
   Multi-layer features are extracted from each decoder branch, concatenated, and fed into two learnable doppelganger classification heads. 
   Each head generates predictions $(\textrm{pred}_{pq}^v, \textrm{pred}_{qp}^v), v\in\{1,2\}$ (where $pq$ and $qp$ denote the symmetrized image pair), supervised by cross-entropy loss. \textbf{(Right)} We use multi-layer decoder features and a Transformer-based classifier head for doppelganger prediction.}
   \label{fig:head_model}
   \vspace{-10pt}
\end{figure*}

\subsection{Improved Doppelganger Classifier}
\label{subsec:imrpoved_classifier}
Recent advances in data-driven models~\cite{Wang2023DUSt3RG3, Leroy2024GroundingIM} have demonstrated impressive results in geometric vision tasks.
In our work, we leverage the multi-level, 3D-aware features extracted from the MASt3R model~\cite{Leroy2024GroundingIM} to train a doppelganger classifier on labeled image pairs, 
as demonstrated in Fig.~\ref{fig:head_model}.

Equipped with large-scale training data and a ViT backbone, the MASt3R model captures an image representation that encodes rich 3D geometric information between paired images. 
However, as MASt3R was originally trained to detect correspondences and similarities, its point maps often conflate doppelganger pairs, yielding incorrect point clouds and poses for these challenging pairs (examples in Fig.~\ref{fig:mast3r_sfm_result}). 
Despite this limitation, we find that the internal features learned by MASt3R contain sufficient information for doppelganger classification without the need to fine-tune MASt3R’s weights.
Moreover, given that the amount of available training data for the doppelganger task is dwarfed by the amount of data used to train MASt3R,
we opt \emph{not} to fine-tune the entire model, 
but instead repurpose these internal features by training an additional output head to predict a doppelganger classification score. 
In Sec.~\ref{subsec:ablation}, we empirically demonstrate that this choice achieves comparable or better visual disambiguation performance, especially on out-of-domain scenes.

\medskip
\noindent \textbf{Multi-level decoder features.}
For an image pair $(I_p, I_q)$, we extract multi-level decoder features from MASt3R's decoder branches.
MASt3R takes two images as input, and uses two intertwined decoders $\text{DecBlk}^1$ and $\text{DecBlk}^2$ to decode encoded features $\mathcal{H}^1=\text{Enc}^1(I_p)$ and $\mathcal{H}^2=\text{Enc}^2(I_q)$.
Each decoder has $B$ blocks, and attends to tokens from the other branch:
\begin{equation}
\small
  f_i^1 = \text{DecBlk}^1_{i}\left(f_{i-1}^1, f_{i-1}^2\right), \\
  f_i^2 = \text{DecBlk}^2_{i}\left(f_{i-1}^2, f_{i-1}^1\right),
\end{equation}
where $f_i^v$ denotes the output tokens from the $i$-th block of the $v$-th branch.
These two branches exchange information to capture the spatial relationships between viewpoints and the global 3D structure of the scene.
For each branch $v$, we concatenate the encoder feature $\mathcal{H}^v$ and the tokens from the decoder blocks into $\mathcal{F}^v=\mathrm{concat}(\mathcal{H}^v, \{f_i^v\}^{B-1}_{i=0})$,
which captures rich, multi-level spatial correspondence information between image pairs.

\medskip
\noindent \textbf{Transformer-based classification heads.}
MASt3R treats one image as the reference frame and projects the other image into that reference frame.
A consequence of this design is that the reconstructed 3D scene for input pair $(I_p, I_q)$ is distinct (or at least, must be in a different coordinate frame) from that of $(I_q, I_p)$. 
Inspired by the asymmetric design, we propose to
1) use separate Transformer heads: We introduce two independent, Transformer-based classification heads $\text{Head}_{dopp}^1$ and $\text{Head}_{dopp}^2$ to process $\mathcal{F}^1$ and $\mathcal{F}^2$ respectively; and
2) symmetrize image pairs such that the model evaluates both $(I_p, I_q)$ and $(I_q, I_p)$ to decide whether the given pair of images are true match or not.
Thus, we end up with four scores for each image pair:
\begin{equation}
    \text{pred}^v_u = \text{Head}_\mathrm{dopp}^v(\mathcal{F}^v_u), u \in\{pq,qp\}, v \in \{1,2\}.
\end{equation}
Essentially, the two heads serve as expert evaluators, each examining how likely the image pair is to be a true match (or doppelganger) from different aspects,
and switching the input order allows the model to analyze the spatial relationships from both directions. 
We empirically show the effectiveness of this design in Sec.~\ref{subsec:ablation}.
Both $\text{Head}_{dopp}^1$ and $\text{Head}_{dopp}^2$ are supervised by the cross-entropy loss,
encouraging $\mathcal{S} = \{\text{pred}^v_u\}$ to all give high probabilities for positive matches and low probabilities for negative ones.

\medskip
\noindent \textbf{Test-time voting.}
At test time, we combine the four scores $\mathcal{S}$ via a voting mechanism to make a final decision.
Specifically, 
if the majority of the heads predict that the pair is a positive match,
we take $\max{(\mathcal{S})}$ as the final score for the image pair. 
Conversely, if the majority of the heads vote for a negative match, 
$\min{(\mathcal{S})}$ is used as the final score. 
Otherwise, we average the scores from the four heads.
This approach ensures that the final decision reflects the strongest evidence supporting the consensus among the heads and thereby improves the reliability of the classifier.
Eq.~\ref{eq:vote_mechanism} elaborates on this voting mechanism:
\begin{equation}
\footnotesize
    S_{\text{final}} = 
    \begin{cases} 
       \max(\mathcal{S}) & \text{if } \sum_{s \in \mathcal{S}} \mathbf{1}(s > 0.5) > \sum_{s \in \mathcal{S}} \mathbf{1}(s < 0.5), \\
       \min(\mathcal{S}) & \text{if } \sum_{s \in \mathcal{S}} \mathbf{1}(s > 0.5) < \sum_{s \in \mathcal{S}} \mathbf{1}(s < 0.5), \\
       \mean(\mathcal{S}) & \text{if } \sum_{s \in \mathcal{S}} \mathbf{1}(s > 0.5) = \sum_{s \in \mathcal{S}} \mathbf{1}(s < 0.5).
    \end{cases}
\label{eq:vote_mechanism}
\end{equation}

\subsection{Evaluating Doppelganger correction in SfM}
\label{subsec:dopp_evaluation}
There is currently no reliable benchmarking method for evaluating the accuracy and correctness of SfM reconstructions specifically in terms of how well they address the doppelganger issue.
Prior work~\cite{Cai2023DoppelgangersLT} relied on manual inspection of each out model to assess the effectiveness of their approach---a time-consuming, unquantifiable, and potentially error-prone process. 
In our work, we propose a benchmarking method for qualitatively evaluating 
reconstructed models with respect to doppelgangers. 
We leverage mapping sites like \href{https://www.mapillary.com/app/}{Mapillary},
which provide images with location metadata that can serve as probes for validating a 3D model.
Note that our method targets common scenarios where geotags are unavailable, or so noisy as to not be useful. Therefore, we do not consider the use of geotags for the reconstruction task itself---we explore pure visual disambiguation---but instead gather them from specialized sources for ground truth evaluation.

\begin{figure}[t]
  \centering
   \includegraphics[width=.9\linewidth]{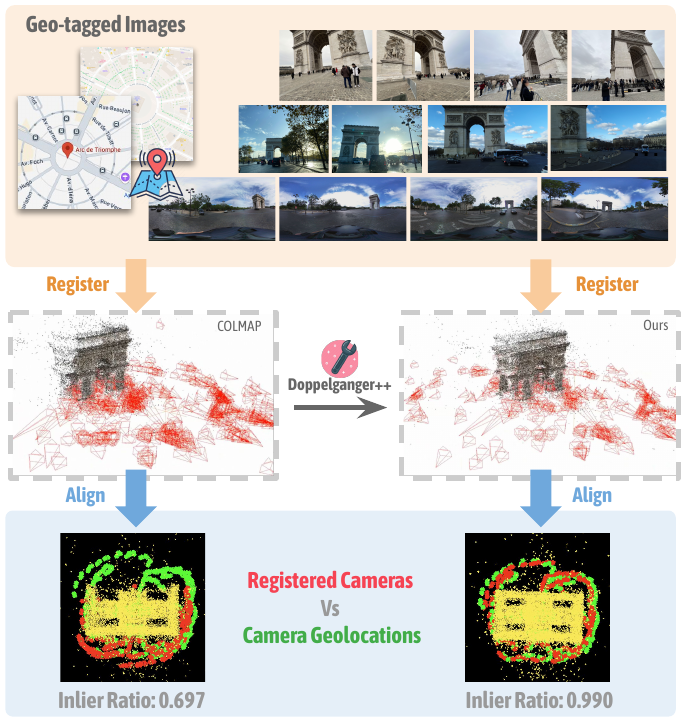}
   \vspace{-5pt}
   \caption{\textbf{Evaluation of doppelganger correction in SfM.} \textbf{(Top)} We first collect sequences of geo-tagged Mapillary images around the target location and register them to the SfM model. 
   Then, we use RANSAC to align the registered cameras and their geolocations. The inlier ratio is computed as an indicator of model accuracy. \textbf{(Bottom)} In the model corrupted by doppelganger pairs, the registered cameras all collapse to one side. We see that the camera poses estimated with COLMAP (right, in red) do not align well with the geotags (green), leading to a low inlier ratio, but our method leads to a much closer alignment.}
   \label{fig:geo_tag_prob}
   \vspace{-15pt}
\end{figure}

Specifically, for a given scene with known geolocation, we acquire from Mapillary a set of nearby geo-tagged images and register them to the reconstructed model with COLMAP~\cite{schonberger2016structure}. 
We then apply RANSAC to robustly estimate a similarity transformation between the registered camera positions and their corresponding image geolocation metadata (converted to ECEF coordinates).
We use the resulting RANSAC \emph{inlier ratio} as an indicator of the model’s correctness. 
As an example, in Fig.~\ref{fig:geo_tag_prob}, we can observe that before correction, the registered cameras and their geolocations show significant misalignment, 
with the cameras collapsing to one side due to similar-looking surfaces. 
After correction, the registered cameras align closely to their true geolocations, indicating that doppelganger pairs are correctly separated in the model.
If the model is split into multiple components, we calculate the weighted average of inliers ratios among the components:
\begin{equation}
\text{IR} = \sum_{i=1}^{N} \left(\frac{I_i}{T_i} \cdot \frac{T_i}{\sum_{j=1}^{N} T_j}\right) = \frac{\sum_{i=1}^{N} I_i}{\sum_{i=1}^{N} T_i}, 
\end{equation}
where $I_i$ is the number of RANSAC inliers and $T_i$ is the number of registered Mapillary images in the $i$-th component.
While alignment error could also serve as a metric, we opt to use the inlier ratio to reduce sensitivity to geolocation inaccuracies, 
as image geolocations may not always be precise.
This strategy can also be used to detect broken reconstructions from datasets like MegaScenes~\cite{Tung2024MegaScenesSV}, which contains over $100$K SfM results from Internet photos around the world.

\section{Experiments}
\label{sec:experiments}
\subsection{Experimental setup}
\noindent \textbf{Dataset.}
Our training set is comprised of the Doppelganger dataset~\cite{Cai2023DoppelgangersLT} (dubbed DG) and the new \visym dataset.
Specifically, the {\dgdata} training set includes 73K image pairs, nearly evenly split between positive and negative pairs. 
From \visym, we use 47K image pairs across 26 sites as training data, evenly split into positive and negative pairs. 

We evaluate on two tasks, 1) pairwise image visual disambiguation and 2) scene reconstruction by integrating our classifier into COLMAP~\cite{schonberger2016structure}. 
For pairwise image visual disambiguation, 
we utilize the {\dgdata} test set, and mined an additional 3,180 pairs from 7 {Visym} sites (distinct from the 26 training sites), equally divided into positive and negative pairs.
Additionally, we created a test set from the Mapillary Street-Level Sequences dataset~\cite{Warburg2020MapillarySS}.
This Mapillary dataset spans diverse urban and suburban environments and captures a wide range of seasons, weather conditions, cameras, lighting, and structural settings, with each image geo-tagged by GPS and compass angle. 
Using a similar filtering approach to the \visym dataset, we mined 1,500 positive and 1,500 negative pairs as an out-of-domain test set.

For the scene reconstruction task, we evaluate on 16 scenes sampled from Heinly~\etal~\cite{heinly2014correcting}, Wilson~\etal~\cite{wilson2013network}, 
MegaScenes~\cite{Tung2024MegaScenesSV} and 5 \visym test scenes.
These scenes are challenging for conventional SfM pipelines as well as prior disambiguation methods due to subtle differences between distinct surfaces and repetitive patterns.

\medskip
\noindent \textbf{Metrics.}
For pairwise visual disambiguation evaluation, we report Average Precision (AP) and ROC AUC scores, following the protocol in~\cite{Cai2023DoppelgangersLT}. 
Additionally, 
we report precision at a fixed recall and recall at a fixed precision.
These statistics help characterize a model’s trade-off between achieving high precision and potentially missing true positives.

To comprehensively evaluate SfM reconstructions, we report the number of images in the SfM results, and the geo-alignment inlier ratio described in Sec.~\ref{subsec:dopp_evaluation}.
Respectively, more images in the SfM model imply less false pair pruning,~\ie, better class separation; 
and a higher geo-alignment inlier ratio suggests that the reconstructed model is more likely to be accurate and complete.
Note that the inlier ratio is also influenced by the accuracy of the geo-tagged images collected from Mapillary.
Thus, if one test scene has more accurate geo-tagged images than another, it is likely to yield a higher inlier ratio. 
Therefore, rather than comparing inlier ratios across different scenes, this metric is more useful for evaluating different reconstruction methods on the same scene, as the same set of geo-tagged images is used.

\medskip
\noindent \textbf{Baselines.}
Our primary baseline is~\cite{Cai2023DoppelgangersLT}, as their learning-based approach consistently outperforms heuristic-based methods like thresholding on the number of post-RANSAC inliers~(\eg SIFT~\cite{lowe2004distinctive}+RANSAC~\cite{Fischler1981RandomSC}, D2-Net~\cite{Dusmanu2019D2NetAT}+RANSAC, SuperPoint~\cite{detone2018superpoint}+SuperGlue~\cite{Sarlin2019SuperGlueLF}) for pairwise visual disambiguation; and is also superior than previous approaches~\cite{heinly2014correcting, wilson2013network, Cui2015GlobalSB, yan2017distinguishing} for SfM disambiguation.

\subsection{Implementation}
\noindent \textbf{Model details.}
Our doppelganger classifier head $\text{Head}_\textrm{dopp}$ is implemented with a Transformer encoder comprised of 3 layers, each with input and output dimension of 768. 
Each layer has 8 attention heads and a two-layer feed-forward network with a hidden dimension of 2048. 
Pre-layer normalization and residual connections are applied within each layer.
The transformed tokens are aggregated via max pooling and linearly projected into $\text{pred}^v$.
We freeze the weights of the public MASt3R~\cite{Leroy2024GroundingIM} model and only train the two doppelganger classification heads,
supervised by a cross-entropy loss.
We train for 5 epochs with a batch size of 8 using Adam~\cite{Kingma2014AdamAM} with a learning rate of $1\times 10^{-4}$. More details can be found in the supplementary.

\medskip
\noindent \textbf{Integration with SfM.}
Following~\cite{Cai2023DoppelgangersLT}, we integrate Doppelgangers++ into SfM to enhance its disambiguation ability.
SfM takes a collection of images $\mathcal{I}=\{I_i\}^n_{i=0}$ and generates geometrically verified image pairs $\mathcal{P}=\{(I_p, I_q)\}$, forming a scene graph $\mathcal{G}=(\mathcal{I}, \mathcal{P})$ with images as nodes and pairs as edges.
Using $\mathcal{G}$, SfM computes camera poses and reconstructs a 3D point cloud.
Our doppelganger classifier acts as a filter on edges in $\mathcal{G}$,
removing pairs below a probability threshold $\tau$ to eliminate spurious matches due to repeated or symmetric structures.
We integrate Doppelgangers++ into both COLMAP~\cite{schonberger2016structure} and MASt3R-SfM~\cite{Duisterhof2024MASt3RSfMAF} to showcase the effectiveness of the method.
Notably, with COLMAP, SIFT features are computed for scene graph construction and MASt3R features for pruning; 
whereas with MASt3R-SfM, MASt3R features serve for both scene graph construction and pruning, forming a more efficient pipeline.

\begin{figure*}
  \centering
   \includegraphics[width=\linewidth]{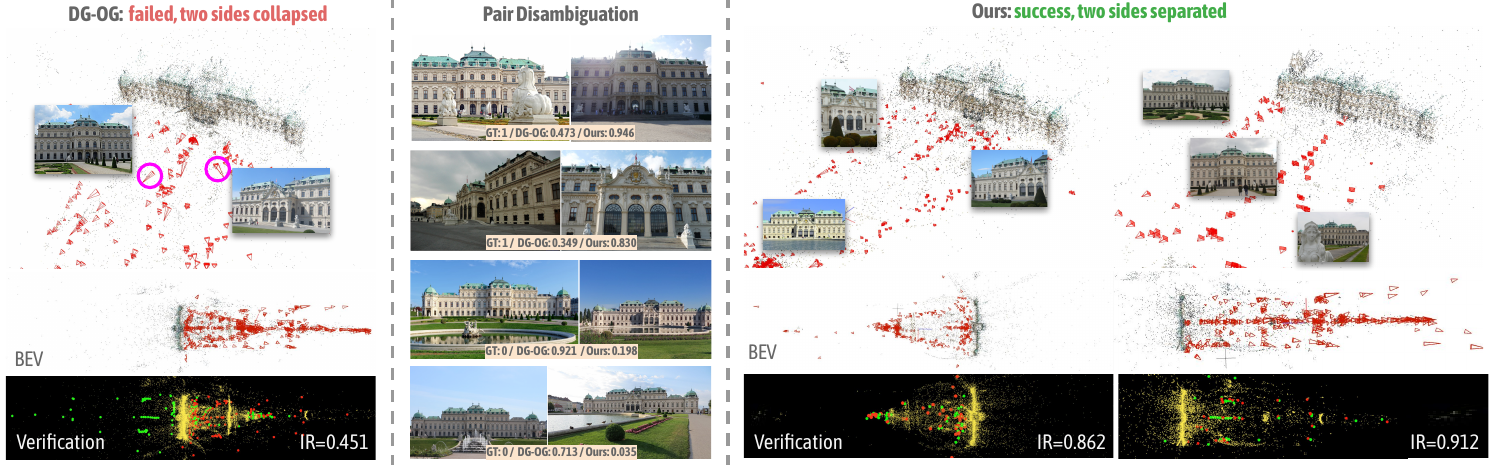}
   \vspace{-20pt}
   \caption{\textbf{SfM Disambiguation on MegaScenes~\cite{Tung2024MegaScenesSV}.} \textbf{(White background)} SfM results from DG-OG~\cite{Cai2023DoppelgangersLT} and ours. \textbf{(Black background)} Verification using geo-tagged images, \textcolor{red}{red} points represent registered cameras and \textcolor{green}{green} points represent geolocations, inlier ratio (IR) is labeled on the bottom right. DG-OG fails to disambiguate this scene, predicting incorrect scores for image pairs. Our method correctly splits the model into two clean components.}
   \label{fig:megascene_sfm_vis}
   \vspace{-15pt}
\end{figure*}

\begin{figure}
  \centering
   \includegraphics[width=\linewidth]{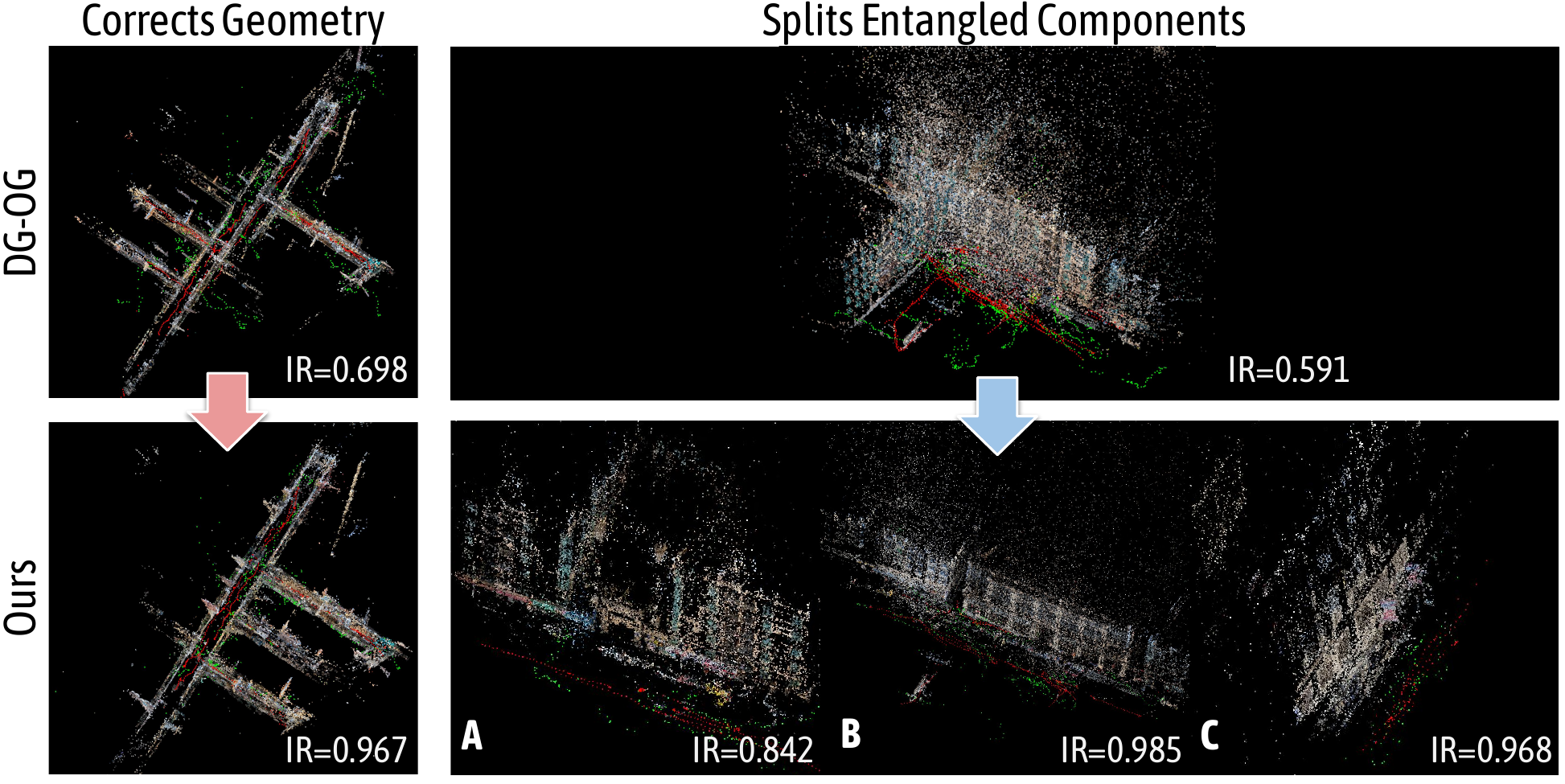}
   \vspace{-20pt}
   \caption{\textbf{SfM disambiguation on \visym.} We show that our classifier is more robust than DG-OG~\cite{Cai2023DoppelgangersLT} on test scenes from new domains, like everyday street scenes. DG-OG has difficulty disambiguating such scenes, leading to incorrect geometry.}
   \label{fig:visym_sfm_vis}
   \vspace{-5pt}
\end{figure}

\subsection{Pairwise Visual Disambiguation}
We compare our approach with the original Doppelganger work~\cite{Cai2023DoppelgangersLT} (dubbed \emph{DG-OG}) under two experimental settings: 
1) we use the same training data as in~\cite{Cai2023DoppelgangersLT},~\ie the {\dgdata} training set, to train our model. 
Both models are evaluated on three benchmark test sets: {\dgdata}, {\visym}, and {Mapillary}. 
Under this configuration, the {\dgdata} test set is in-domain, while {\visym} and {Mapillary} are out-of-domain scenarios;
2) we expand the training data by including the new \visym training set and retrain both DG-OG and our model.
Evaluation is conducted on the same three test sets, with only the {Mapillary} test set being out-of-domain this time.

Quantitative results are in Tab.~\ref{tab:classical_vs_ours}.
Under the first setting, where both methods are trained on {\dgdata}, 
our model shows clear improvements across all three test sets.
Specifically, on the in-domain {\dgdata} test set, 
our model achieves higher AP and ROC AUC, 
indicating that it maintains high precision across all recall levels,
while also being less sensitive to the decision threshold $\tau$ with better class separation. 
On the out-of-domain test sets ({\visym} and {Mapillary}), 
we observe $25\%$ to $65\%$ improvements in both AP and ROC AUC, highlighting the generalizability of our method.

Although our model demonstrates improved performance even when trained solely on the {\dgdata} dataset, 
its precision outside of the training domain is suboptimal for practical use in SfM, where high precision is essential. 
In the second setting, we expand the training set by including \visym data. 
Both methods maintain similar performance on the {\dgdata} test set,
with our model experiencing a slight drop in recall when precision is set to $0.99$. 
On the \visym test set, both models show improvements across all metrics, with ours reaching $99\%$ in both AP and ROC AUC. 
Notably, adding \visym to the training set does not enhance DG-OG's performance on the out-of-domain {Mapillary} test set, 
while ours continues to benefit from increased training diversity. 
These results indicate that the prior approach struggles to generalize effectively, whereas ours shows greater generalization with more varied training data.

\begin{table}[]
\centering
\resizebox{\linewidth}{!}{
\begin{tabular}{cccccc}
\toprule
\multirow{2}{*}{\textbf{Test Data}} & \multirow{2}{*}{\textbf{Method}} & \multicolumn{4}{c}{\textbf{Metrics} (trained on \dgdata\ / trained on \dgdata + \visym)} \\
\cmidrule{3-6}
 &  & AP$\uparrow$ & ROC AUC$\uparrow$ & Prec@Recall=0.85\(\uparrow\) & Recall@Prec=0.99\(\uparrow\) \\
\midrule
\multirow{2}{*}{\centering \textbf{\dgdata}} 
    & DG-OG & 0.954 / 0.956 & 0.944 / 0.947 & 0.901 / 0.910 & 0.611 / 0.614 \\
    & Ours & 0.980 / 0.981 & 0.981 / 0.981 & 0.972 / 0.982 & 0.690 / 0.642 \\
\midrule
\multirow{2}{*}{\centering \textbf{\visym}} 
    & DG-OG & 0.816 / 0.938 & 0.726 / 0.921 & 0.498 / 0.831 & 0.340 / 0.623 \\
    & Ours & 0.936 / 0.991 & 0.909 / 0.990 & 0.892 / 0.999 & 0.542 / 0.901 \\
\midrule
\multirow{2}{*}{\centering \textbf{Mapillary}} 
    & DG-OG & 0.566 / 0.692 & 0.581 / 0.701 & 0.523 / 0.572 & 0.003 / 0.000 \\
    & Ours & 0.950 / 0.968 & 0.944 / 0.958 & 0.927 / 0.942 & 0.310 / 0.736 \\
\bottomrule
\end{tabular}}
\vspace{-5pt}
\caption{\textbf{Evaluation of pairwise disambiguation.} We evaluate DG-OG~\cite{Cai2023DoppelgangersLT} and our method trained on \dgdata~\cite{Cai2023DoppelgangersLT} only and \dgdata+\visym (two numbers per cell) on three test sets.
Both DG-OG and ours benefit from dataset expansion, whereas ours gained more generalizability on out-of-domain test set (Mapillary) after training on both. 
Our classifier constantly demonstrates better precision, recall across all settings.}
\label{tab:classical_vs_ours}
\vspace{-15pt}
\end{table}

\subsection{Structure from Motion disambiguation}
We integrate our classifier trained on \dgdata and \visym into COLMAP’s SfM pipeline~\cite{schonberger2016structure}, 
and evaluate its performance on reconstructing scenes with duplicated and symmetric structures. 
We compare the results with vanilla COLMAP and Cai~\etal~\cite{Cai2023DoppelgangersLT} (DG-OG).
Quantitative results are presented in Tab.~\ref{tab:landmark_evaluation}. 
Our approach registers more images across all scenes than DG-OG and operates without threshold tuning. 
In contrast, DG-OG relies on scene-specific thresholds, such as $\tau=0.97$ for the Church on Spilled Blood
to correctly separate both sides of the church, and $\tau=0.6$ for Ponte di Rialto~\cite{Tung2024MegaScenesSV} to avoid over-segmentation and maintain completeness. 
Notably, DG-OG fails to disambiguate Belvedere (Vienna),
while our method succeeds with a consistent threshold ($\tau=0.8$ across all scenes). 
Our approach also consistently achieves a higher inlier ratio than the baselines, indicating greater accuracy and completeness in the reconstructed models.
As an example, we qualitatively show the results of reconstructing Belvedere (Vienna) in Fig.~\ref{fig:megascene_sfm_vis}, along with verification results using geo-tagged images.

We also evaluate on scenes from the \visym test set. 
Since \visym images come with geolocation metadata, we do not need additional Mapillary images; 
instead, we directly compute the inlier ratio between SfM camera positions and the geolocation metadata after RANSAC. 
Results show that our model effectively prunes spurious pairs and improves the reconstructed model quality, whereas DG-OG sometimes fails to distinguish doppelganger pairs. 
Fig.~\ref{fig:visym_sfm_vis} shows examples where DG-OG encounters difficulties.

\begin{figure}
  \centering
   \includegraphics[width=\linewidth]{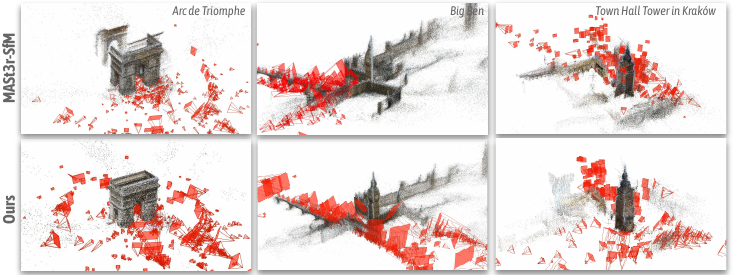}
   \vspace{-20pt}
   \caption{\textbf{MASt3R-SfM w/ Doppelgangers++.} MASt3R-SfM also suffers from doppelganger issues. Our classifier effectively prunes false positive pairs and correctly reconstructs challenging scenes. }
   \vspace{-5pt}
   \label{fig:mast3r_sfm_result}
\end{figure}

In Fig.~\ref{fig:mast3r_sfm_result}, we show that while MASt3R's features are powerful, MASt3R-SfM is not free from doppelganger issues. 
Although our classifier was trained on image pairs mined through COLMAP's feature matching module~(\ie w/ SIFT features), it effectively prunes incorrect matches generated by MASt3R, restoring accurate reconstruction results.

\begin{table}[]
\centering
\resizebox{\linewidth}{!}{
\begin{tabular}{ccccccc}
\toprule
                           & \multicolumn{3}{c}{\textbf{\# SfM-registered Images}} & \multicolumn{3}{c}{\textbf{Inlier Ratio~(Sec.~\ref{subsec:dopp_evaluation})}} \\
\textbf{Test Scenes}          & {COLMAP} & {DG-OG} & {Ours} & {COLMAP} & {DG-OG} & {Ours} \\
\midrule
Alexander Nevsky Cathedral~\cite{heinly2014correcting} & 447          & 444          & 447           & 0.565      & 1.0           & 1.0      \\
Arc de Triomphe~\cite{heinly2014correcting}            & 424         & 384          & 423           & 0.697      & 0.966         & 0.990    \\
Berliner Dom~\cite{heinly2014correcting}               & 1603          & 1588         & 1606          & 0.709      & 0.987         & 0.992    \\
Big Ben~\cite{heinly2014correcting}                    & 398          & 379          & 394           & 0.564      & 0.827         & 0.831    \\
Church on Spilled Blood~\cite{heinly2014correcting}    & 273           & 84+94~(\textcolor{red}{$\tau$=0.97})        & 157+106    & 0.542      & 0.881         & 0.962    \\
Radcliffe Camera~\cite{heinly2014correcting}           & 281          & 91+84        & 94+186        & 0.495      & 0.955         & 0.970    \\
Seville~\cite{wilson2013network}                    & 1498          & 585+272+515      & 615+303+552       & 0.450      & 0.772         & 0.854    \\
Brandenburg Gate~\cite{Tung2024MegaScenesSV}           & 2137          & 1361+570     & 1398+603      & 0.440      & 0.900         & 0.909    \\
Palacio de Comunicaciones~\cite{Tung2024MegaScenesSV}  & 727          & 307+80~(\textcolor{red}{$\tau$=0.6})       & 308+84        & 0.229      & 0.823         & 0.934    \\
Ponte di Rialto~\cite{Tung2024MegaScenesSV}            & 652          & 538+101~(\textcolor{red}{$\tau$=0.6})      & 540+107       & 0.627      & 0.834         & 0.844    \\
York Minster~\cite{Tung2024MegaScenesSV}               & 636          & 200+362      & 206+284       & 0.727      & 0.858         & 0.901    \\
Town Hall Tower, Kraków~\cite{Tung2024MegaScenesSV}  & 298          & 255          & 280           & 0.609      & 0.731         & 0.838    \\
Belvedere (Vienna)~\cite{Tung2024MegaScenesSV}         & 1038          & 851 (\textcolor{red}{fail})       & 457+500       & 0.521      & 0.451         & 0.874    \\
Reichstag (building)~\cite{Tung2024MegaScenesSV}       & 1504         & 997+310      & 1024+356      & 0.469      & 0.804         & 0.862    \\
St. Vitus Cathedral~\cite{Tung2024MegaScenesSV}        & 752          & 673          & 692           & 0.853      & 0.909         & 0.933    \\
Royal Liver Building~\cite{Tung2024MegaScenesSV}       & 212          & 171          & 180           & 0.7        & 1.0           & 1.0      \\
\midrule
VisymSite0010 & 1544 & 520+290 & 1446 & 0.770 & 0.820 & 0.913 \\
VisymSite0023 & 849 & 471+81 & 566+82 & 0.867 & 0.848 & 0.942 \\
VisymSite0028 & 450 & 238+179 & 267+120 & 0.818 & 0.711 & 0.909 \\
VisymSite0042 & 540& 206+207 & 467& 0.863 & 0.924 & 0.963 \\
VisymSite0109 & 1245 & 237+458+78 & 239+612+127 & 0.857 & 0.862 & 0.927 \\

\bottomrule
\end{tabular}}
\caption{\textbf{Evaluation of SfM results}. We compare reconstructions from COLMAP~\cite{schonberger2016structure}, DG-OG~\cite{Cai2023DoppelgangersLT}, and our method. 
$\tau$=0.8 is used unless otherwise stated. The `$+$' indicates split reconstruction components;~\eg, DG-OG and our method split the Radcliffe Camera reconstruction into two components. Because \visym scenes are large, we report statistics on the largest reconstruction component produced by COLMAP, and identify the corresponding (split) components in the results of DG-OG and ours. }
\label{tab:landmark_evaluation}
\vspace{-20pt}
\end{table}

\subsection{Ablation}
\label{subsec:ablation}
\begin{figure}[t]
  \centering
   \includegraphics[width=\linewidth]{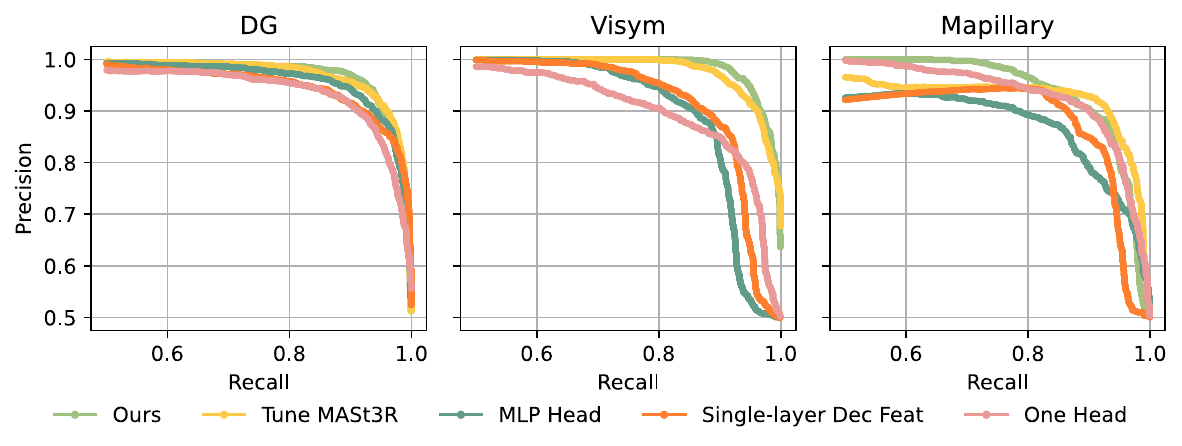}
   \vspace{-20pt}
   \caption{\textbf{Precision-Recall curves of ablation studies.} Metrics are evaluated on models trained with \dgdata and \visym. Full tables can be found in the supplementary.}
   \label{fig:abl_curves}
   \vspace{-15pt}
\end{figure}

In this section, we study the effectiveness of our designs from the following aspects: 
1) fine-tuning the entire model (vs.\ only new heads), 
2) one classification head (vs.\ two separate heads),
3) the architecture of our classification head, and
4) final-layer decoder features (vs.\ multi-layer decoder features).
We evaluate models trained on \dgdata and \visym datasets and show PR curves in Fig.~\ref{fig:abl_curves} on the three test sets with respect to the pairwise classification task. 
A full ablation table can be found in the supplementary.

\medskip
\noindent \textbf{Fine-tuning MASt3R.}
As discussed in Sec.\ref{subsec:imrpoved_classifier}, we choose to train a lighter output head on top of MASt3R features rather than fine-tuning the entire model weights. 
Curves in Fig.~\ref{fig:abl_curves} show that training only the head achieves comparable performance to full model fine-tuning. 
The drop in recall with full model fine-tuning suggests potential overfitting to training data,
making the model more conservative and less generalizable to unseen inputs. 
This tendency also holds on models trained solely on \dgdata data: on the \visym test set, our approach achieved 0.03 higher AP than fine-tuning. On Mapillary test set, ours achieved comparable AP but with 0.2 higher recall at precision 0.99.

\medskip
\noindent \textbf{Single classification head.}
Our design uses two classification heads to process features from the two branches separately.
An alternative approach would be to combine the features from both decoder branches and use a single head to predict the doppelganger score. 
As shown in Fig.~\ref{fig:abl_curves}, this alternative degrades performance consistently.
This may be due to the asymmetric nature of the MASt3R design, where each branch captures different kinds of information that are better to be processed separately.
On the other hand, our voting mechanism between the two heads helps enhance class distinction, outputting lower scores for negative samples and higher scores for positive samples.

\medskip
\noindent \textbf{Head architecture.}
While the decoder features of MASt3R retain rich geometric 3D information, designing a classifier with sufficient capacity to effectively leverage this information for the doppelganger prediction task is non-trivial. 
We compare the performance of our Transformer-based classifier with that of an MLP. 
Fig.~\ref{fig:abl_curves} shows that our Transformer model outperforms the MLP across all metrics and test sets.

\medskip
\noindent \textbf{Single-layer decoder feature.}
Our classifier head takes multi-layer decoder features as input. 
Here we substitute with only the final-layer decoder feature.
Results in Fig.~\ref{fig:abl_curves} show that the multi-layer decoder features are consistently superior to the single-layer ones,
because the classifier is able to analyze doppelganger factors from different aspects and levels. 
See supplementary for more detailed analysis.

\section{Conclusion}
\label{sec:conclusion}
We propose Doppelgangers++ as an effective approach to handling visual aliasing in 3D reconstruction. 
We introduce a new \visym dataset, featuring 
images from diverse daily scenes, and develop rules to mine doppelganger data from the dataset to enrich our training data.
We train a Transformer-based classifier that leverages geometric 3D features to classify image pairs with high precision and recall across various scenes. 
Doppelgangers++ integrates seamlessly into existing SfM pipelines, enhancing reconstruction accuracy without extensive parameter tuning. 
Further, we propose a validation method using geo-tagged map images, offering a more comprehensive and automatic way to assess SfM accuracy and model completeness. 
Extensive experiments show that Doppelgangers++ significantly improves visual disambiguation and 3D reconstruction quality in complex scenes.

\section{Acknowledgment}
We thank Brandon Li and Joseph Tung for their valuable discussion and help with the webviewer. 
This work was supported by the Intelligence Advanced Research Projects Activity (IARPA) via Department of Interior/Interior Business Center (DOI/IBC) contract number 140D0423C0035. The U.S. Government is authorized to reproduce and distribute reprints for governmental purposes notwithstanding any copyright annotation thereon. Disclaimer: The views and conclusions contained herein are those of the authors and should not be interpreted as necessarily representing the official policies or endorsements, either expressed or implied, of IARPA, DOI/IBC, or the U.S. Government.
This research was also supported in part by the National Science Foundation under award IIS-2212084.

{
    \small
    \bibliographystyle{ieeenat_fullname}
    \bibliography{main}
}



\end{document}